# An investigation towards wavelet based optimization of automatic image registration techniques

Arun P.V[1]                    S.K. Katiyar

Department of Civil
MANIT-Bhopal, India

## Abstract

Image registration is the process of transforming different sets of data into one coordinate system and is required for various remote sensing applications like change detection, image fusion, and other related areas. The effect of increased relief displacement, requirement of more control points, and increased data volume are the challenges associated with the registration of high resolution image data. The objective of this research work is to study the most efficient techniques and to investigate the extent of improvement achievable by enhancing them with Wavelet transform. The SIFT feature based method uses the Eigen value for extracting thousands of key points based on scale invariant features and these feature points when further enhanced by the wavelet transform yields the best results.

## 1. Introduction

Image registration is the process of geometrically aligning images of the same scene taken from different viewpoints at different times or by different sensors. It is a fundamental image processing technique and is important for integrating information from different sensors, finding changes in images taken at different times, inferring three-dimensional information from stereo images, and recognizing model-based objects. High-resolution images have introduced new challenges for traditional processing methods, including current image registration techniques, for the following reasons as increased relief displacement, requirement of more control points, and increased data volume. In order for spatial resolution to become smaller than 1m, the altitude of the sensors is being decreased, which increases the relief displacement and causes localized distortion related to landscape height. Precisely locating control points in high-resolution images is not as simple as with moderate-resolution and to obtain precise registration, a large number of control points must be manually selected across the whole image, which is a tedious and time consuming job. High data volume often affects the processing speed in the image registration and hence automation in this context is highly demanding.

---

[1] arunpv2601@gmail.com

Literature revels a great deal of automation approaches towards registration in context of medical images as well as remote sensing images and hence only important works have been considered in this study. Schenk et.al (2001) suggested that the feature based methods are more preferred over intensity methods since the derived features are inherently unique and similarity is based on the attributes which are comparatively invariant to transformations. Lucas (1984) used a kernel density estimation scheme for approximating smoothed probability density function (PDF) from a point set. Jian et.al (2005) adopted density estimation with Gaussian kernels to mitigate the effects of outlier and noise over previous approaches. Wang et al. (2009) generalized the Kullback-Leibler (KL) divergence measure to the Jensen-Shannon (JS) divergence measure for point set registration where as the use of Density Power Divergence (DPD) as an estimator between pairs of PDFs was proposed by Basu et al (2008). The above discussed point registration techniques provide a scope of improvement using wavelets for point optimization and matching.

Li et al (1997) designed a fast image registration algorithm based-on randomized contour matching and Zhaou et.al (2003) used the curves to register remote sensing image with the rigid transformation. Maintz et al (1998) tried to register images using elastic contours and internal points, but he just used contours and points consecutively instead of simultaneously. Wen peng et.al (2008) proposed a new curve-based local image registration method, which can elastically register the images in the frame work of feature curve which is approximated to a B-spline for matching purpose. However, all the algorithms as above only make use of one kind of feature information, points, contours or curves, and an attempt of combination of them is neglected. These methods can be further enhanced using the Contourlet transform for extracting orientation preserved curves. Stefan et.al (2007) provided a comparison of eight optimization methods for non rigid registration based on the maximization of mutual information, in combination with a deformation field parameterized by cubic B-splines. Malviya et al (2009) had suggested computationally and qualitatively efficient method that uses the Haar wavelet transform along with mutual information for image-registration. MI based approaches can be improved using wavelet for optimizing the gradient matching and maximizing the mutual information between voxels.

SIFT transform convert images into a large collection of feature vectors that are invariant to transformation, partially invariant to illumination changes, and robust to local geometric distortion. The SIFT features extracted from two images are compared based on their location, scale, and orientation to accomplish registration (Robert et.al, 2004). The integration of wavelet and SIFT transforms in feature matching process can provide an improved context sensitivity. The wavelet-based approaches preserve the spectral characteristics of the multi-spectral images better than the standard PCA and HIS methods (Medha, 2009;Malviya, 2009). A hybrid approach based on wavelets, feature extraction technique, normalized cross-correlation matching and relaxation-specific image matching techniques was suggested by Hong et al (2008) and the approach could successfully select enough control points to reduce the local distortions.

Wavelet-Modulus Maxima method suggested by Fonesca et.al (1998) detects control points from the local modulus maxima of the wavelet transform and correlation coefficient is used as a measure of similarity to select the control points through best pair-wise fitting. The wavelet based approaches can be further enhanced based on the context by varying the type of wavelets used as orthogonal, Daubechies, Gabor feature-based methods etc. The polynomial-model image registration is often used to register different images, because it is included in the commercial software and easy to operate. However, for those high-resolution images with terrain relief, the registration accuracy produced by the polynomial-model registration method is poor (Medha et al., 2009). Although a variety of proposed image registration methods can be found in the literature of the past few years, to date, it is still difficult to find an accurate, robust, and automatic image registration method (Min Li, 2006).

We investigate the various efficient methodologies used in the context of image registration and the possibility of their optimization using wavelets. The objective of this research work is to study the improvement achievable for the most efficient automatic image registration techniques by enhancing them with Wavelet transform. This research paper compiles and analyzes the comparative performances of various advanced methods of automatic image registration when enhanced by the wavelet transform. In this paper we have used the DT-CWT form of Haar wavelet along with Gabor feature based methods and contourlets to enhance the existing methods so as to improve their accuracy and reduce the time complexity.

## II. Theoretical background: Wavelet

The wavelet transform has gained a great deal of interest due to its time localization and multi resolution properties (Laine et al., 1994). Saxena et al. (1997) used two kinds of wavelets namely quadratic-spline and Daubechies six-coefficient wavelet to extract small features from electrocardiogram (ECG) signals. Wavelets can be applied for various image processing tasks as denoising (Sayed et al., 2008; Cheng et al., 2006), enhancement (Malviya et al., 2009; Medha,2009;Das,2006), and feature extraction (Su et al., 2005;chen,2011). Fourier transforms (FT) lack time localization as frequency components are attributed to the entire time signal and not to specific parts of it (Daubechies et al., 2007), 2000). Windowed Fourier Transforms (WFT) achieves localization however uses fixed size windows that cannot be adjusted to suite the speed of the changing phenomena observed in the input signals (Das et al., 2006). Wavelets solve this problem by using the so called mother wavelet, which can be scaled and translated to achieve both time localization and multi-resolution (Zhou et al., 2009). Common wavelet transforms can represent an image in three different directions as horizontal, vertical, and diagonal (Daubechies et al., 2007). The three selective directions are good enough for extracting features from an image which can be effectively used for enhancing various image registration techniques (Sayed et al., 2008; Chen, 2011).

The common wavelet transform causes aliasing and exhibits shift variance (Chang et al., 1993) and hence many modified wavelet transforms have been introduced. Kingsbury (Kingsbury, 2000) introduced the complex wavelet transform to effectively solve the two problems of aliasing and shift variance. The DT-CWT has been shown to be suitable for feature based image registration. Hill (Hill et al., 2003) used DT-CWT coefficients to construct the texture gradient map for the watershed segmentation algorithm. The most efficient forms found in literature with regard to image registration were the complex wavelet transform, the tree-structured wavelet transforms (Chen et al., 2011), and the dual-tree complex wavelet transforms (Gonzalez et al., 2002).

## III. Methodology

The various automatic image registration algorithms available in the literature were analyzed and the most efficient methodologies were implemented in Matlab. Scopes of optimization for these techniques at various steps were analyzed and wavelet transform was applied to improve the feature matching capability as well as for key point optimization. Contourlet based enhancement was adopted for point set and curve based registration methods. The Satellite images namely PAN, CARTOSAT, and Google earth images of Bhopal area were used as test images for assessing the accuracy of the optimized algorithm with corresponding existing methods. Prior to registration, in order to make same resolution, high resolution images were degraded to match with coarse spatial resolution data. The various algorithms were implemented in MATLAB and the accuracies were computed using ERDAS imagine and the results were compared. The quality of the registration process was measured using the following criteria: normalized cross- correlation coefficient (NCCC) and root mean square error (RMSE) .The methodology adopted for this research work is as given in (Figure 1).

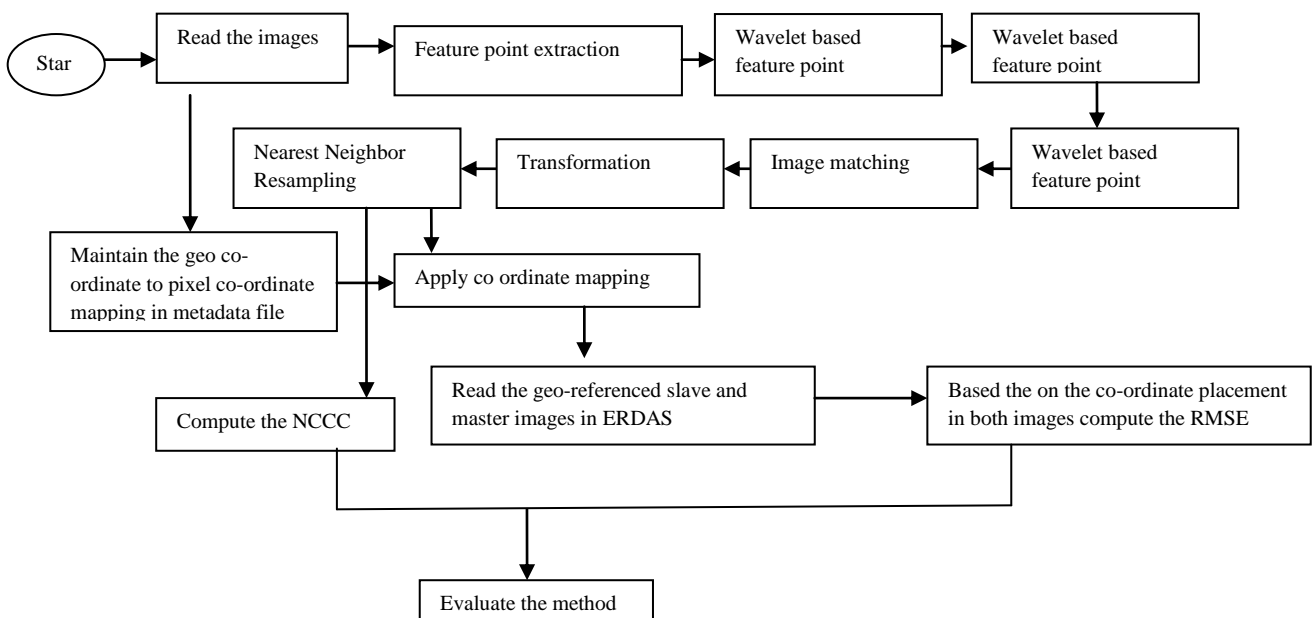

Figure 1. Algorithm for wavelet based optimization

## IV. Data resources and study area

The investigations of present research work have been carried out for the satellite image s of different spatial resolution sensors for the Bhopal city in India and given in the (Table 1). The study area is new market area of Bhopal city having area central point coordinates 23° 55' N Latitude and 76° 57' E Longitude. The algorithms have been implemented in the MATLAB software environment and the accuracy of the techniques was verified using Erdas, Matlab. The Erdas-Imagine v9.1 was also used for the pre-processing of images and other image analysis tasks.

Table 1. Data sources description

| S.NO | Imaging sensor | Resolution(m) | Satellite | Area | Acquisition Date |
|------|----------------|---------------|-----------|------|------------------|
| 1 | CARTOSAT-1 | 2.5 | IRS P5 | Bhopal(India) | 5[th] April 2009 |
| 2 | LISS-III | 24 | IRS P5 | Bhopal(India) | 5[th] April 2009 |
| 3 | LISS-IV | 2.5 | IRS P5 | Bhopal(India) | 16[th] March 2010 |

## V. Results

The various efficient image registration techniques have been implemented in Matlab and have been optimized with wavelet transform. The optimized methodologies were compared with the conventional approach and the accuracy has been verified on various satellite images as Google Earth, LISS 4, and Cartosat. The results of observations are as summarised in (Table 2). Normalized Cross Correlation Coefficient measures the similarity between the images. NCCC value ranges from [0-1] and an NCCC value of unity indicates perfectly registered images. RMSE value indicates the error in registration and a least RMSE value is preferred for a perfect registration. The execution time was also analyzed using MATLAB counter function and generally categorized as high (>1 min), medium (30-60 sec), low (<30 sec).The results from the table are self explanatory that the new method out performs the existing automatic registration methods.

Table 2. Accuracy comparison

| S. N o | TEST DATA | | TECHNIQUE | NCCC | RMSE | NCCC (optimiz ed) | RMSE (optimiz ed) | EXECU TION TIME | EXECU TION TIME | PREFERENCE |
|---|---|---|---|---|---|---|---|---|---|---|
| | Master Image | Slave Image | | | | | | | | |
| 1 | Cartosat | LISS4 | Point set Based | .71 | 1.93 | .81 | 1.02 | High | Low | Contourlet based enhancement |
| 1 | Cartosat | LISS4 | Surface | 0.61 | 3.85 | 0.65 | 3.80 | High | High | Contourlet based enhancement |
| 2 | Cartosat | LISS4 | Curve | 0.59 | 4.82 | 0.61 | 4.02 | High | High | Presence of linear features |
| 3 | Cartosat | LISS4 | Mutual Information (MI) Based | 0.67 | 1.57 | 0.71 | 1.02 | High | Low | Intensity variations are captured and matched |
| 4 | Cartosat | LISS4 | Soft computing methods | 0.63 | 2.5 | 0.64 | 1.85 | High | Low | Prior terrain knowledge |
| 5 | Cartosat | LISS4 | SIFT Feature based methods | 0.87 | 0.61 | 0.93 | 0.56 | High | High | Optimization using Haar transform |

The feature based methodologies are giving optimal results as compared to the area based approaches. The MI technique is appropriate for the multi modal registration however it is having the disadvantages of area based methods and can be improved by wavelet based optimization. The increase in dimensionality increases the computational complexity of these algorithms. The SIFT feature based method uses the Eigen value for extracting thousands of key points based on scale invariant features and these feature points when further enhanced by the wavelet transform yields the best results.

## VI. Conclusions

Investigations of this research work have shown that the wavelet based methods and the SIFT feature based techniques are most accurate methods for the registration of remotely sensed images. But these methods are computationally complex and take more processing time. The investigation revealed that Wavelet optimization techniques can be applied to improve the feature matching capability as well as for key point optimization. The SIFT feature based method uses the Eigen value for extracting thousands of key points based on scale invariant features and these feature points when further enhanced by the wavelet transform yields the best results.

## VII. References

Audette M. A., "An algorithmic overview of surface registration techniques for medical imaging," *Medical image Analysis*, vol.4, no.1, pp.201–217, 2004.

Basu A., Harris I. R., Hjort N. L., and Jones M. C., "Robust and efficient estimation by minimising a density power divergence," *Biometrika*, vol.85, no.3, pp.549–559, 2008.

Chen Q., "Image registration and its applications in medical image," PhD thesis, Vrije University, Brussels, Belgium, 2011.

Cheng H., Zheng N., and Sun C., "Boosted Gabor features applied to vehicle detection," 18th International Conference on Pattern Recognition, pp.662-666, 2006.

Chi Kin Chow, "Surface Registration using a dynamic genetic algorithm," *Journal of Pattern Recognition*, vol.37, no.1, pp.105-117, 2004.

Das D., Singh N. K., and Sinha A. K., "A comparison of Fourier transform and wavelet transform methods for detection and classification of faults on transmission lines," *Power India Conference*, *IEEE ,* vol., no., pp.7, 15th Dec 2006.

Daubechies I., Teschke G., and Vese L., "Iteratively solving linear inverse problems under general convex constraints," *Inverse Problems Imag.,* vol.1, no.1, pp.29–46, 2007.

Elsen P. A. V. D., "Medical image matching: a review with classification," *IEEE Engineering in medicine and biology,* vol.12, no.1, pp.26-39, 1998.

Ezzeldeen R. M., "Comparative study for image registration techniques of remote sensing images," *The Egyptian Journal of Remote Sensing and Space Science*, vol.13, no.6, pp.31- 36, 2010.

Flusser J., "A moment-based approach to registration of images with affine geometric distortion," IEEE *Transaction on Geoscience and Remote Sensing*, vol.32, no.1, pp.382–387, 1994.

Fonesca L. M., and Max H. M., "Automatic Regestration of Satellite Images," *Proceedings on IEEE transaction of computer society*, vol.23, no.2, pp 219-226, 1997.

Gonzalez R.C., and Woods R.E., Digitial Image processing, Prentice Hall,Upper Saddle River, NJ, 2nd edition,2002.

Hill R. C., Canagarajah C.N., and Bull D.R., "Image segmentation using a texture gradient based water shed transform," IEEE Transactions on image processing, vol.12,no.12,pp.1618-1633,2003.

Hong G., and Zhang Y., "Wavelet-Based Image Registration Technique for High Resolution Remote Sensing Images," *Computers & Geosciences*, vol.34, no.11, pp.1708-1720, 2008.

Jian B., and Vemuri BC., "A robust algorithm for point set image registration using mixture of gausians," *Tenth IEEE International Conference on* computer *Vision,* vol.2, pp.1246-125, 2005.

Kingsbury N. G., "Complex wavelets for shift invariant analysis and filtering of signals," *Journal of Applied and Computational Harmonic Analysis*, vol.10, no.3, pp.234–253, 2001.

Laine A., Fan J. A., and Yang W. H., "Wavelets for contrast enhancement of digital mammography," *IEEE Engineering Medical Biology Magazine,* vol.14, no.5, pp.536–550, 1995.

Li M., and Vitanyi P., "An Introduction to Kolmogorov Complexity and Its Applications," *Springer- Verlag*, vol.46, no.3, pp.130-135, 1997.

Lowe P and David G., "Distinctive image features from scale-invariant key points," *International Journal of Computer Vision,* vol.60, no.2, pp.91-110, 2004.

Lucas B. D., "Generalized image matching by the method of differences," PhD thesis, School of Computer Science, Carnegie–Mellon University, Pittsburgh, PA, pp.120-130, 1984.

Maes F., Collignon A., Vandermeulen D., Marchal G., and Suetens P., "Multimodality image registration by maximization of mutual information," *IEEE Transactions on Medical Imaging,* vol.16, no.2, pp.187-198, 1997.

Maintz J. B. A., and Max A. V., "A Survey of Medical Image Registration," *Medical Image Analysis,* vol.2, no.1, pp.1-36, 1998.

Malviya A., and Bhirud S.G., "Wavelet based image registration using mutual information," Emerging *Trends in Electronic and Photonic Devices & Systems*, *ELECTRO '09. International Conference on*, vol., no., pp.241-244, 22-24 Aug 2009.

Medha V. W., "Image Registration Techniques: An overview," *International Journal of Signal Processing, Image Processing and Pattern Recognition,* vol.2, no.3, pp.1-5, 2009.

Min Li., "A region-based multi-sensor image fusion scheme using pulse-coupled neural network," *Journal of Pattern Recognition Letters*, vol.27, no.16, pp.1948–1956, 2006.

Min Xu., "A Subspace Method for Fourier-Based Image Registration," *IEEE Geoscience and Remote Sensing Letters*, vol.6, no.3, pp.491-494, 2009.

Peng Wen, "Medical Image Registration Based-on Points, Contour and Curves," *International Conference on Biomedical Engineering and Informatics*, vol.55, no.3, pp.132-136, 12th April 2008.

Saxena S. C., Sharma A., and Chaudhary S. C., "Data compression and feature extraction of ECG signals," *International Journal of Systems Science*, vol.28, no.5, pp.483-498, 1997.

Sayadi O., and Shamsollahi M.B., "Model - Based fiducial points extraction for baseline wander ECGs," IEEE Transaction on biomedical engineering, vol.55, no.1, pp.347-351, January 2008.

Schenk H., "Correlation of projection radiographs in radiation therapy using open curve segments and points," *Journal of Medical Physics*, vol.19, no.2, pp.329-334, 2001.

Stefan Klein., "Medical Image Registration Using Mutual Information and B-Splines," *IEEE Transactions on Image Processing*, vol.16, no.12, pp.2879-2890, 2007.

Wang H., Yang Y., Ma S., and Guo C., "Automatic Object Extraction Based on Fuzzy Mask," *Intelligent Systems and Applications, ISA International Workshop on*, vol.23, no.1, pp.1-4, May 2009.

Zhou S., Tang B., and Chen R., "Comparison between Non-stationary Signals Fast Fourier Transform and Wavelet Analysis," *Intelligent Interaction and Affective Computing, ASIA'09,* vol.22, no.6, pp.128-129, June 23[rd], 2009.

Zitov´a B., and Flusser J., "Image registration methods: a survey," *Image and Vision computing,* vol.21, no.11, pp.977-1000, 2003.